\documentclass[conference]{IEEEtran}
\IEEEoverridecommandlockouts
\usepackage{cite}
\usepackage{amsmath,amssymb,amsfonts}
\usepackage{algorithmic}
\usepackage{graphicx}
\usepackage{textcomp}
\usepackage{xcolor}
\usepackage{multirow}
\usepackage{booktabs}
\usepackage{amssymb}
\usepackage{amsmath}
\usepackage{graphicx}

\def\BibTeX{{\rm B\kern-.05em{\sc i\kern-.025em b}\kern-.08em
		T\kern-.1667em\lower.7ex\hbox{E}\kern-.125emX}}

\begin{document}
\title{Trust Oriented Explainable AI \\ for Fake News Detection}
	
\author{
	\IEEEauthorblockN{1\textsuperscript{st} Krzysztof Siwek}
	\IEEEauthorblockA{\textit{Warsaw University of Technology} 
		\\ Warsaw, Poland\\
		 ORCID: 0000-0003-2642-2319 \\
		 krzysztof.siwek@pw.edu.pl}
	\and
	\IEEEauthorblockN{2\textsuperscript{nd} Daniel Stankowski}
	\IEEEauthorblockA{\textit{Student of} \\
		\textit{Warsaw University of Technology}\\
		 Warsaw, Poland }
	\and
	\IEEEauthorblockN{3\textsuperscript{rd} Maciej Stodolski}
	\IEEEauthorblockA{\textit{Warsaw University of Technology} \\
		Warsaw, Poland \\
		ORCID: 0009-0003-0644-8105 \\
		maciej.stodolski@pw.edu.pl}
}
			
\maketitle

\begin{abstract}
This article examines the application of Explainable Artificial Intelligence (XAI) in NLP based fake news detection and compares selected interpretability methods. The work outlines key aspects of disinformation, neural network architectures, and XAI techniques, with a focus on SHAP, LIME, and Integrated Gradients. In the experimental study, classification models were implemented and interpreted using these methods. The results show that XAI enhances model transparency and interpretability while maintaining high detection accuracy. Each method provides distinct explanatory value: SHAP offers detailed local attributions, LIME provides simple and intuitive explanations, and Integrated Gradients performs efficiently with convolutional models. The study also highlights limitations such as computational cost and sensitivity to parameterization. Overall, the findings demonstrate that integrating XAI with NLP is an effective approach to improving the reliability and trustworthiness of fake news detection systems.
\end{abstract}


\section{Introduction}
This work investigates and compares selected Explainable Artificial Intelligence (XAI) methods in the context of fake news detection. We briefly present XAI methodologies, covering post-hoc and ante-hoc explainability, local and global interpretation strategies, and techniques such as SHAP, LIME, Anchors, Integrated Gradients, and attention mechanisms.

In the experimental section, we implement a complete pipeline for fake news detection, encompassing data preparation, classification model development, and integration of XAI techniques, specifically SHAP, LIME, and Integrated Gradients. The results confirm the research hypothesis that the application of XAI enhances the transparency and interpretability of the NLP models while preserving high classification performance. Although quantitative differences between XAI methods remain moderate within the same model, each technique provides unique interpretive value: SHAP yields detailed and faithful local explanations, LIME offers accessible and intuitive interpretations, and Integrated Gradients perform particularly well for convolution based architectures, combining high quality attributions with relatively low computational cost. Visual analyses further highlight the alignment between explanation patterns and the intrinsic properties of CNN and LSTM models.

Our work also identifies limitations associated with XAI, including computational overhead, sensitivity to parameterization, and the risk of misinterpretation by end users. A comparative, model specific evaluation emerges as the safest approach, avoiding problematic generalizations across different architectures. 

Overall, the our research demonstrates that combining XAI with NLP provides an effective and forward looking strategy for improving the credibility, validation, and practical utility of systems designed for fake news detection. The findings offer a foundation for future work involving transformer based architectures, rule based explanations, end user studies, model calibration, and linguistically grounded perturbation protocols.

\section{Related Work}
The spread of digital misinformation has forced the deployment of high capacity deep learning (DL) models for automated veracity assessment. While architectures such as Transformers and Graph Neural Networks (GNNs) achieve high empirical accuracy, their "black box" nature presents significant barriers to deployment in banking and public health contexts \cite{AB2023106087_7}. The lack of transparency in these models limits error auditing and does not provide the evidentiary support required by human fact checkers. Explainable Artificial Intelligence has emerged as a critical corrective, shifting the research focus from mere predictive performance to the generation of human understandable rationales that justify classification decisions.

Current research in explainable fake news classification is splited into model agnostic post-hoc interpretations and intrinsically interpretable architectures.

The dominant paradigm utilizes post-hoc techniques to interpret pretrained neural classifiers. Local Interpretable Model agnostic Explanations (LIME) and SHapley Additive exPlanations (SHAP) remain the foundational tools for identifying token level saliency \cite{NIPS2017_8a20a862},  \cite{art3}. These methods quantify the contribution of specific input features, typically words or phrases to the final veracity label. SHAP utilizes game theory to quantify the individual contribution of each input feature to the model's final prediction. By calculating the Shapley value, the method determines how much a specific feature (such as a keyword or metadata attribute) shifts the prediction away from the average baseline, ensuring a fair and mathematically grounded distribution of influence among all variables. \cite{NIPS2017_8a20a862}.

While effective for local interpretability, these methods primarily offer feature attribution rather than causal narrative, often failing to explain the linguistic nuances or logical fallacies inherent in deceptive content.

To circumvent the limitations of post-hoc approximations, recent efforts have focused on transparent models. The Tsetlin Machine represents a significant advancement in this area, utilizing propositional logic to form interpretable conjunctive clauses \cite{bhattarai2021explainabletsetlinmachineframework_4}. Unlike the opaque weights of a neural network, Tsetlin Machine's provide explicit logical rationales for why a news item is flagged. Furthermore, hybrid architectures attempt to fuse the representative power of DL with interpretable modules, such as FastText integrated deep learning frameworks, to balance classification robustness with structural transparency \cite{art3}.

A pivotal transition in the state of the art is the movement from feature level attribution to Generative Justification. Leveraging the capabilities of Large Language Models (LLMs), researchers have introduced frameworks such as LLM-GAN and defense based evidence extraction \cite{wang2024llmganconstructgenerativeadversarial_6}. These systems do not merely highlight suspicious keywords; they synthesize natural language rationales that explain the discrepancy between a claim and verified evidence. This approach transforms XAI from a diagnostic tool for developers into a communicative tool for end users, though it introduces risks related to hallucinated explanations and the potential for the model to justify a correct prediction with incorrect reasoning.

Fake news detection has evolved beyond isolated textual analysis to incorporate the social aspect of misinformation. Current models now integrate Social Propagation Patterns, utilizing GNNs to explain veracity based on how information spreads through network topologies \cite{amri2023exfakeexplainablefakenews_5} and Incorporate Source Reputation and account behavioral cues into the explanation module \cite{Kim03072019}, \cite{SANNA2025106090}.

The latest area in XAI is multimodal fusion which explaining the interplay between inflammatory and unbelievable text, and manipulated visual media. This is an area that remains underdeveloped but is essential for addressing sophisticated disinformation campaigns \cite{AB2023106087_7}.

There is a mismatch between the statistical significance of features and human understanding. Features identified as important by the model may not exactly match those considered important by humans \cite{martens2025bewareexplanationsai}.

\section{Mechanisms and Manifestations}
Fake news refers to intentionally fabricated or misleading information presented in the form of legitimate news. Although misinformation has existed for centuries, the digital era has dramatically accelerated its production and reach. Social media platforms, online news aggregators, and algorithmic recommendation systems enable false content to spread rapidly, often outpacing verified information. Fake news draws its persuasive power from emotional appeal, simplicity of narrative, and the tendency of individuals to favor information consistent with their preexisting beliefs.

The dissemination of false information as a tool for social, political, or military influence has a long history. However, it was only the development of the internet, and particularly social media, that led to an unprecedented acceleration of this phenomenon. Fake news, or false or partially manipulated information stylized as reliable journalistic content, has begun to play a key role in shaping public opinion. The effectiveness of fake news stems not only from its content, but also from its form, method of distribution, and the social context in which it is received. 

The consequences of fake news extend far beyond individual beliefs. At the societal level, it undermines trust in institutions, weakens democratic discourse, and fuels polarization. In critical contexts, such as public health, elections, or international conflict, disinformation can cause tangible harm by influencing behavior, distorting political processes, or escalating tensions. Recognizing these risks, researchers emphasize the need for effective verification mechanisms, including factchecking initiatives, media literacy education, and technological tools capable of identifying and explaining misleading content.

\section{Neural Networks in NLP}
Natural Language Processing (NLP) is a subfield of computer science and artificial intelligence focused on enabling machines to understand and process human language. Its importance stems from the contextual nature of human generated data only by capturing linguistic and semantic context can computers interpret and use such information effectively. 

Contemporary NLP uses a variety of neural architectures, including convolutional neural networks (CNNs), recurrent neural networks (RNNs), graph neural networks (GNNs), and attention based models such as transformers. Their advantage lies in replacing manual feature design with dense, low dimensional vector representations that encode syntactic and semantic properties of language. These representations are learned directly from data, making NLP systems more adaptable and effective across tasks.

\subsection{Recurrent Networks}
The first neural language models were created using simple feedforward neural networks, which took a sequence of words of a fixed length as input and predicted the next word in the sequence. Due to their limited context, these models were only able to analyze short text fragments, which limited their usefulness in tasks requiring the capture of long term semantic relationships.

Recurrent Neural Networks (RNNs) introduced feedback loops that allow information from previous time steps to influence future computations, enabling the preservation of context in longer sequences. Their training relies on Backpropagation Through Time (BPTT), which propagates errors across the entire sequence. However, classic RNNs suffer from the vanishing gradient problem: repeated matrix multiplications over long sequences cause gradients to shrink, preventing the model from learning long range dependencies. As a result, RNNs perform well on short sequences but struggle with longer texts.

To solve the problem of vanishing gradients, in \cite{lstm_Hochreiter} was proposed the Long Short Term Memory (LSTM) architecture. The LSTM utilizes a cell state (memory gate) and three distinct gates that manage information through sigmoid and tanh activation functions. This gate determines which information from the previous cell state is no longer relevant and should be discarded. It processes the previous hidden state and the current input, outputting a value between 0 (completely discard) and 1 (completely keep). The input gate decides which new information will be stored in the cell state. It works in two parts: a sigmoid layer decides which values to update, and a tanh layer creates a vector of new candidate values to be added to the state. The output gate determines what the next hidden state should be. This hidden state contains information on previous inputs and is used for the actual prediction. It filters the current cell state, ensuring that only the relevant parts of the memory are passed forward to the next time step.

Another significant modification of RNNs are Gated Recurrent Units (GRUs) \cite{chung2014empiricalevaluationgatedrecurrent}. GRUs are a simplified alternative to LSTMs, combining some gate functions and eliminating the need for a separate memory cell. They also learn faster than LSTMs while maintaining comparable quality.

In practice, both LSTM and GRU have contributed significantly to the advancement of language models, enabling more efficient modeling of sequences and capturing long term dependencies in text.

\subsection{Convolutional Neural Networks}
Unlike sequential models (such as LSTMs), which process inputs chronologically and thus suffer from a lack of computational parallelization, Convolutional Neural Networks (CNNs) allow for simultaneous operations across multiple input positions. While CNNs were initially developed for computer vision, they have been successfully adapted for Natural Language Processing (NLP) by applying one-dimensional filters over word embeddings.

A notable advancement in this domain is the Gated Convolutional Neural Network (GCNN) \cite{HEINRICH2021113494}. In this architecture, convolutional layers are augmented with a Gated Linear Unit (GLU).

\subsection{Transformers}
A major breakthrough in modern language modeling is the introduction of Transformer architectures \cite{vaswani2023attentionneed}. They have an encoder–decoder structure and make use of attention mechanisms, and incorporate several innovations that greatly improve performance in tasks like machine translation.

In a Transformer, the input consists of a sequence of tokens encoded as vector representations known as input embeddings. To these embeddings, additional vectors indicating each token’s position in the sequence, positional encodings, are added. The encoder then processes all token representations at the same time, rather than sequentially, enabling highly efficient parallel computation and more effective modeling of long range dependencies.

The Transformer architecture is characterized by a symmetric stack of $N$ identical layers, designed to optimize both contextual representation and sequential generation.

Each encoder layer is composed of two primary sublayers: Multi-Head Self-Attention (MHSA) and a position-wise Feed-Forward Network (FFN). The self-attention mechanism enables the model to compute a weighted representation of each token by attending to all other tokens in the sequence. By calculating Query (Q), Key (K), and Value (V) matrices, the model assigns attention scores that define the contextual relevance of distant tokens.To capture diverse linguistic relationships, Multi-Head Attention executes this process across several independent heads or subspace representations simultaneously. The resulting outputs are concatenated and linearly transformed. Following this, the FFN processes each token representation independently and identically, typically consisting of two linear transformations with a ReLU activation in between.

The decoder generates the output sequence step by step and consists of N layers with three sublayers: Masked Multi Head Self Attention, Encoder Decoder Attention, and a Feed Forward network. The encoderdecoder attention mechanism lets the decoder focus on relevant parts of the encoder’s representations when producing each token.

After passing through all decoder layers, the resulting vector is projected to vocabulary size via a linear layer and converted into word probabilities using Softmax. Generation proceeds sequentially: once a token is produced, it is reinserted into the first Masked SelfAttention layer as an output embedding, while masking ensures the model cannot access future, unseen tokens.

\section{XAI as a tool for interpreting NLP models}
A key distinction in Explainable AI methods concerns their scope of applicability. Techniques are divided into modelspecific methods, which rely on the internal structure of a particular model type, and model agnostic methods, which treat the model as a black box and can be applied to any architecture.

Model specific approaches, such as gradientbased techniques in neural networks, attention visualizations in Transformers, or TreeSHAP for decision trees offer high precision and lower computational cost but lack universality. They cannot be easily transferred between different model families.

Model agnostic methods, including LIME, KernelSHAP, Partial Dependence Plots, ICE, and Anchors, are more flexible and work with any predictive model, making them suitable for environments that use diverse algorithms or require consistent explanations. Their drawbacks include higher computational cost and potentially less faithful interpretations.

In practice, model agnostic methods are valuable in regulated domains like healthcare, finance, or automotive systems, where transparent explanations are mandatory. Model specific methods are preferred when analyzing a single architecture in depth and when efficient, finegrained explanations are needed.

\subsection{SHAP - Shapley Additive Explanations}
SHAP, Shapley Additive Explanations, is a method for interpreting individual predictions of machine learning models \cite{SHAP_1}, \cite{SHAP_2}. It is based on Shapley values from cooperative game theory, which provide a theoretically sound way to attribute a model’s output to its input features.

In SHAP, each feature value for a given instance is treated as a player in a cooperative game, and the model prediction is the total payout. Shapley values quantify how much each feature contributes to the prediction relative to a baseline, typically the average model output. The final prediction is expressed as the sum of the baseline value and the individual feature contributions, ensuring an additive and interpretable explanation.

Exact computation of Shapley values is computationally expensive because the number of possible feature coalitions grows exponentially with the number of features. Therefore, practical implementations rely on efficient approximations or model specific solutions, such as TreeSHAP for tree based models. SHAP is widely applied to explain black box models in both classification and regression tasks.

\subsection{LIME - Local Interpretable Model agnostic Explanations}
LIME (Local Interpretable Model agnostic Explanations) is a local, model agnostic approach to explaining black box models \cite{LIME}. Its objective is to approximate the behavior of a complex model around a specific instance using a simpler and interpretable surrogate model.

The central idea is to treat the original model as an unknown function that can be queried for predictions. To explain a single prediction, LIME generates perturbed versions of the input instance and obtains corresponding outputs from the black box model. These synthetic samples form a new local dataset around the instance of interest.

An interpretable model, such as linear regression or a decision tree, is then trained on this local dataset. During training, samples are weighted according to their proximity to the original instance, ensuring that the surrogate model focuses on faithfully approximating the decision boundary in the local neighborhood. The resulting explanation reflects which features were most influential for that specific prediction.

Because it is model agnostic and locally focused, LIME is widely used to interpret complex models in classification and regression tasks, particularly in text and image analysis.

\subsection{Integrated Gradients Method}
Integrated Gradients (IG) is an attribution method for explaining predictions of deep neural networks within an axiomatic framework \cite{IG_publ}. IG is a path attribution method. Instead of evaluating the gradient of the prediction function at a single input point, which may be unstable or locally uninformative, it integrates gradients along a continuous path from a baseline input to the actual instance. The baseline is typically a reference point such as a zero vector or a black image.
	
Formally, feature attributions are computed as the integral of the model’s gradient with respect to each input feature along the straight line connecting the baseline and the input instance. This procedure measures the average marginal effect of each feature on the model output across the entire path.
	
As a result, Integrated Gradients provide a principled and mathematically consistent way to quantify which input features contributed most strongly to a specific prediction.

\subsection{Risk of Misinterpretation}
Explainable AI aims to make complex models more transparent by providing systematic and interpretable accounts of their decisions. In practice, however, explanations may be misleading or misinterpreted, even when they are technically correct, especially explanations of black box models are particularly vulnerable to reliability concerns and misuse \cite{EHSAN2024100971}, \cite{Miscund_XAI}, \cite{VANROYEN2026112013}.
	
A primary risk is the illusion of causality. Explanations often reflect correlations learned from training data rather than true domain level causes. For example, a vision model may associate a specific background color with a class label if such correlation is present in the dataset. 
	
Another limitation is instability. Different XAI methods can produce divergent explanations for the same prediction, and even small perturbations of the input may substantially change results. This variability undermines robustness and may lead users to selectively accept explanations that confirm prior expectations.
	
Explanations are also susceptible to manipulation. It has been demonstrated that models can be intentionally modified to generate misleading explanations while preserving biased or undesirable decision rules. Such practices, sometimes referred to as fairwashing, create the appearance of fairness without altering the underlying behavior of the system.
	
Finally, human cognitive biases introduce additional risk. Users may develop overconfidence based on simplified explanations, experience confirmation bias, or generalize local explanations beyond their valid scope. The so called illusion of explanatory depth can further reinforce unwarranted trust in the system.
	
For these reasons, XAI outputs should be treated as analytical aids rather than definitive evidence. In high stakes applications, inherently interpretable models may offer a more reliable alternative than relying exclusively on post hoc explanations of opaque systems.

\section{Analysis of Fake News Detection Models}

\subsection{Data acquisition and processing}
The experiments were conducted using the publicly available ISOT Fake News Dataset \cite{database_fake_news}, published on the Kaggle platform. It contains news articles divided into two categories: true news and fake news (Table \ref{tab1}). Articles classified as true were obtained from Reuters.com, while fake news was collected from websites deemed unreliable and flagged as sources of disinformation by PolitiFact and Wikipedia. The dataset comprises texts published primarily in 2016-2017, focusing on politics and international news.

\begin{table}[ht]
	\centering
	\caption{Division of the ISOT set into classes and thematic categories (subjects).}\label{tab1}
		
	\begin{tabular}{ l c l c }
		\hline
		\textbf{Class (News)} & \textbf{No of articles} &
		\textbf{Category (subject)} & \textbf{No of articles} \\ \hline
		
		\multirow{2}{*}{Real news} 
		& \multirow{2}{*}{21417} 
		& World news & 10145 \\ \cline{3-4}
		& & Politics news & 11272 \\  \hline
		
		\multirow{8}{*}{Fake news}
		& \multirow{8}{*}{23481}
		& World news & 10145 \\ \cline{3-4}
		& & Politics news & 11272 \\ \cline{3-4}
		& & Government news & 1570 \\ \cline{3-4}
		& & Middle-east & 778 \\ \cline{3-4}
		& & US news & 783 \\ \cline{3-4}
		& & Left news & 4459 \\ \cline{3-4}
		& & Politics & 6841 \\ \cline{3-4}
		& & News & 9050 \\ \hline
		
	\end{tabular}
\end{table}

\subsection{Feature Engineering and Text Vectorization}
Despite the high quality of the dataset and its popularity in scientific research, preliminary analysis revealed a number of technical issues that were resolved before further use of the data using standard, routine text data preprocessing methods.

After initial data cleaning, it was necessary to transform the article texts into a numerical representation that would enable their use in machine learning models. For this purpose, a classic approach based on token sequences and vectorization using the TensorFlow Keras library was employed.

In the first step, a dictionary was built based on all articles in the collection, limiting the number of words included to the 20,000 most frequently occurring ones. This restriction allows for data dimensionality reduction and eliminates rare tokens that do not contribute significant information to the classification process. Additionally, a special <OOV> (out of vocabulary) token was introduced, which replaces words not included in the dictionary, ensuring greater stability of the vectorization process.

The next step involved converting each document into a sequence of indexes corresponding to subsequent words in the dictionary. To standardize the sample length, all sequences were adjusted to a maximum length of 750 tokens, with shorter texts padded with zeros and longer ones truncated. This parameter was experimentally selected to capture the full context of most articles while avoiding excessively increasing the dimensionality of the input matrices.

As a result, each article was described with a 750-length vector, and the complete dataset was transformed into a numerical matrix that can be directly used in neural network models.

Articles were labeled with binary labels: 0 for false news and 1 for true news. The data were then divided into training, validation, and test sets at a rate of 63\%, 7\%, and 30\%, respectively, allowing for a reliable assessment of the quality of the models built.

\subsection{Neural Network Architecture}
\begin{figure}[ht!]
	\begin{center}
		a) \includegraphics[width=0.2\textwidth]{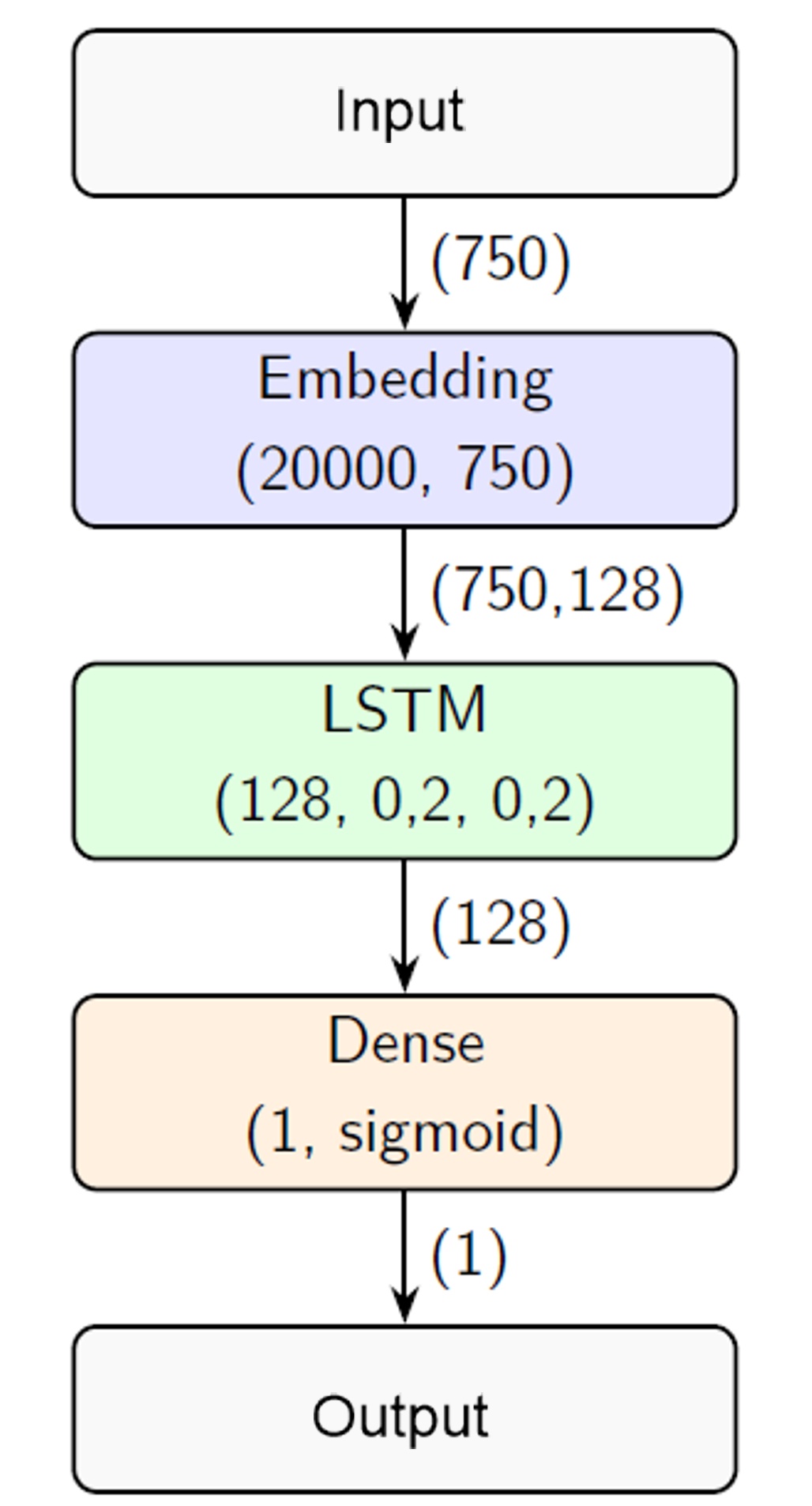}
		b) \includegraphics[width=0.2\textwidth]{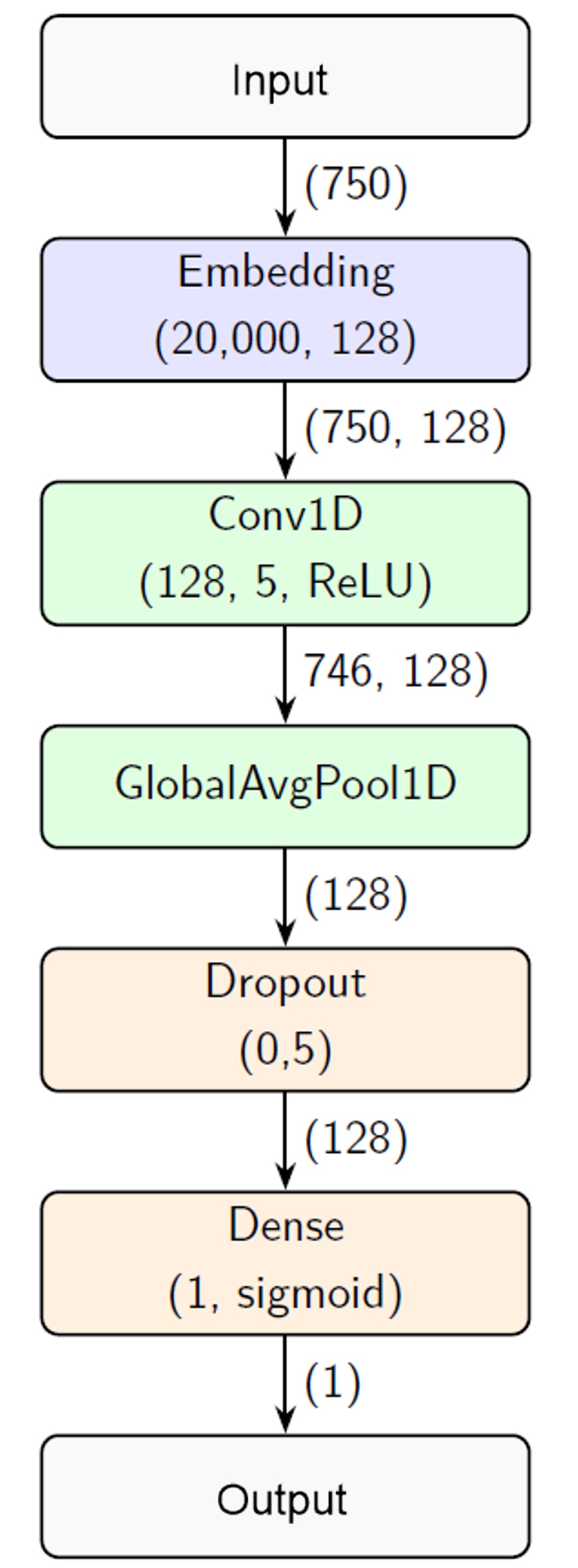}
	\end{center}
	\caption{Model architecture: a) LSTM, b) CNN} \label{fig1_2}
\end{figure}

After initial vectorization, two neural network architectures were evaluated: a sequential LSTM based model and a convolutional alternative (Fig. \ref{fig1_2}).

In the first approach, the input consists of token sequences of length 750. These sequences are passed to an Embedding layer that maps each token to a 128 dimensional vector representation. The embedded sequence is then processed by a single LSTM layer with 128 units, enabling the model to capture long term dependencies within the text. To reduce overfitting, both dropout and recurrent dropout were applied with a rate of 0.2, randomly deactivating connections at the input and recurrent levels during training. Final classification is performed by a Dense layer with one neuron and a sigmoid activation function, producing the probability that a given article belongs to the true news class.

The other model, a convolutional architecture, was implemented to focus on local textual patterns rather than sequential dependencies. After the 128 dimensional Embedding layer, a one dimensional convolutional layer with 128 filters and a kernel size of 5 was used to detect salient n-gram features. This operation reduces the sequence length from 750 to 746 time steps while preserving 128 feature maps. Feature aggregation is performed using \texttt{GlobalAveragePooling1D}, yielding a fixed length representation and providing greater stability in the context of XAI analyses. A Dropout layer with a rate of 0.5 further mitigates overfitting, and final classification is again carried out using a single neuron Dense layer with sigmoid activation.

\subsection{Training and Validation Process}
Both models (LSTM and CNN) were trained using the same framework: the binary crossentropy loss function, the Adam optimizer, and the accuracy metric. A series of experiments were conducted to select the optimal hyperparameters. The data partitioning described earlier (with a fixed random seed \texttt{random\_state=42}) was used; during training, a separate portion of the data was used as the validation set, while the final quality assessment was performed solely on the test set, which was not involved in the training or hyperparameter selection process. The batch size was set to 64.

\subsection{Implementation of XAI methods}
The explanation layer was integrated with the existing input data representation: sequences of token identifiers of fixed length $L=750$ with padding at index 0 (PAD). No retokenization of the test set was performed; the ID → word mapping was performed using a dictionary saved in JSON format (with \texttt{<OOV>} support) during training. The interface to methods expecting text (e.g., SHAP/LIME) was implemented by reconstructing the word sequence solely for the purpose of calling a given library, while explanations from the libraries were transformed into a vector of length L so that each position corresponded to the same position in the input sequence. This ensures that a token at position $j$ in the data has the same evaluation at position $j$ in the vector. PAD positions were omitted both in the explanation presentation and in sequence modification operations; input modification was implemented by PAD substitution, which preserves the sequence's shape and consistency with the model. To ensure repeatability, constant random generator seeds were set.

\subsubsection{SHAP}
A processing procedure was established: a text masker generates multiple versions of a sentence with hidden words, and then SHAP asks the model to evaluate these versions. For this purpose, a simple classification function $f()$ was developed, which accepts a list of sentences, converts them into sequences of identifiers, runs the model, and returns a pair of probabilities $[P(0), P(1)]$ in the format expected by the DeepSHAP library. Based on the obtained ratings, attribution values for words are calculated, which are then assigned to positions in the input sequence. To reduce runtime, for each instance, only 100 partially masked text variants were generated and evaluated by the model.

\subsubsection{LIME}
Similar to SHAP, this method works with strings and uses a function that converts text into sequences of identifiers, runs the model, and returns class probabilities.
LIME creates multiple slightly modified versions of the sentence (e.g., replacing or deleting words) close to the original and, based on this, trains a simple surrogate model describing the local behavior of the classifier.
In practice, approximately 1000 perturbations were generated for each instance, and up to 20 of the most influential words were presented. The returned weights for the words were then mapped to positions in the input sequence of length $L$. This ensures that the LIME output directly matches the representation used by the models.

\subsubsection{Integrated Gradients}
Integrated Gradients has also been implemented as a gradient method operating directly on the embedding representation. The reference point is an "empty" text (a sequence of PADs), and then the change in the model output is observed as one smoothly transitions from this point to the actual input. The transition occurs in the vector space of the Embedding layer.

In the implementation, the model is treated as two parts: the Embedding layer and the Subsequent model layers. Embedding vectors are calculated for the input ($xemb$) and for the reference point ($bemb$). Several dozen points (typically $32-50$) are selected along the line between $bemb$ and $xemb$. For each of them, the gradient of the output value relative to the embeddings is calculated. After averaging the gradients, the result is multiplied by the difference ($xemb-bemb$) and summed over the embedding dimensions. This creates a vector of length $L$, constituting the impact assessment for individual tokens. PAD positions contribute close to zero because the baseline is PAD. The computational cost increases linearly with the number of steps, allowing for a trade off between speed and smoothness. The resulting vector is positionally consistent with the input (length $L$), allowing for direct use in further analyses and visualizations.

\subsection{Explanation Visualization}
A native text graph was used for the visualization of the SHAP method: each token is assigned a color and intensity corresponding to the direction and strength of its contribution to the model's decision (shades of red indicate a shift towards the true class, and shades of blue indicate a shift towards the false class). The entire article is displayed with highlights arranged along the sequence. A natively interactive presentation layer (tooltips, hovering, scrolling) facilitates the analysis of longer texts and the precise reading of local contributions.

\begin{figure*}[h!]
	\begin{center}
		\includegraphics[width=0.8\textwidth]{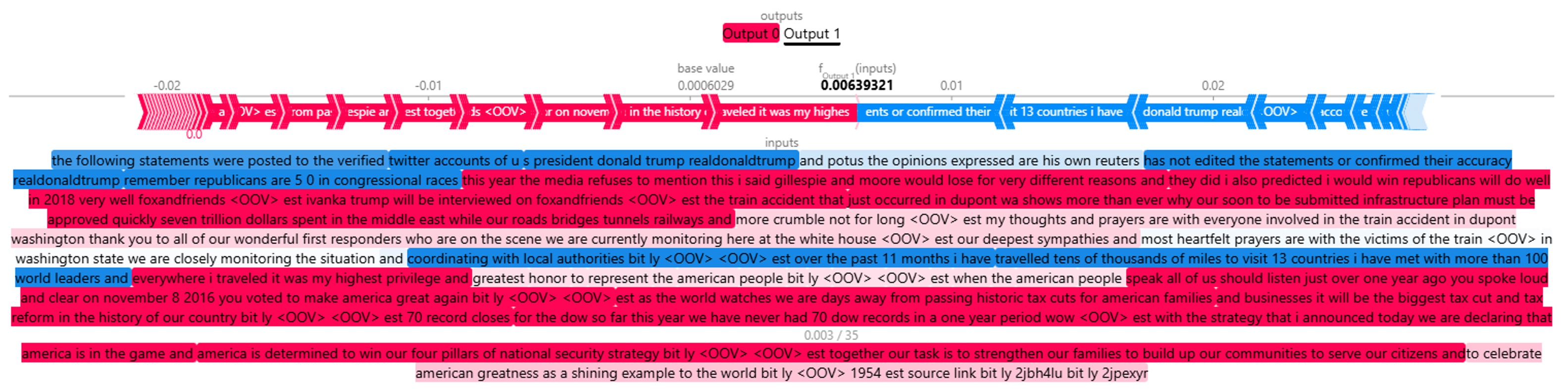}
		
	\end{center}
	\caption{SHAP visualization for CNN model} \label{fig_shap_vis_cnn}
\end{figure*}

The observed saliency maps well reflect the inductive properties of the compared architectures. In the CNN model (Figure \ref{fig_shap_vis_cnn}, attributions cluster into compact, local patches, corresponding to the activation of n-gram filters and aggregation mechanisms. As a result, coherent sentence fragments that push the decision towards a given class are highlighted.

\begin{figure*}[ht]
	\begin{center}
		\includegraphics[width=0.8\textwidth]{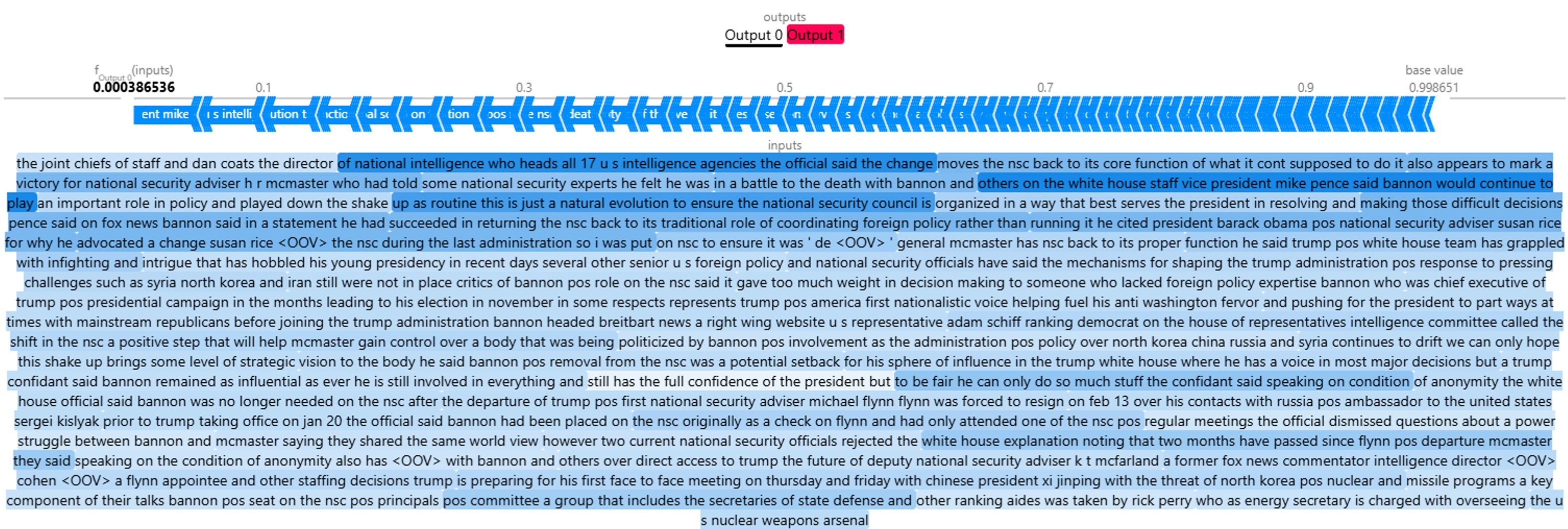}
		
	\end{center}
	\caption{LIME visualization for CNN model} \label{fig_lime_vis_cnn}
\end{figure*}

In the LSTM model, the inputs are more dispersed in time, a consequence of gating
and accumulating information over longer sections of the sequence, resulting in highlights covering wider
areas of text (Figure \ref{fig_lime_vis_cnn}). This difference should be considered an advantage. SHAP, despite its model agnosticity,
faithfully represents the local decision boundary of a specific architecture, thus demonstrating how a given
model actually uses context.

\begin{figure*}[ht]
	\begin{center}
		\includegraphics[width=0.8\textwidth]{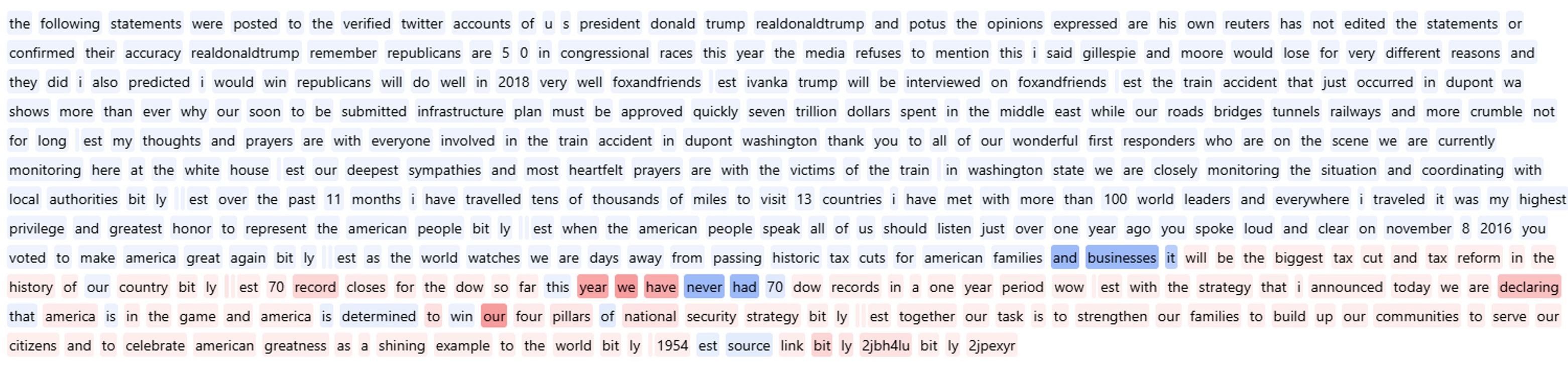}
		
	\end{center}
	\caption{Integrated Gradients visualization for CNN model} \label{fig_ig_vis_cnn}
\end{figure*}

A full text heat map (Figure \ref{fig_ig_vis_cnn}) was prepared for the IG visualization. Each token is colored depending on the sign of its contribution to the decision (shades of red shift the decision toward the true class, shades of blue toward the false class), and the color intensity is proportional to the absolute value of the evaluation. Similar to SHAP, Integrated Gradients reflects the nature of both architectures, but to a less visible extent.

\subsection{Classification Metrics}
Each method (SHAP, LIME, IG) calculates a score vector $a = (a_1, \ldots, a_L)$ for the sequence $x = (t_1, \ldots , t_L)$. To compare results uniformly, the order of importance is determined based on the absolute values of $|a_i|$, and the set $S_k$ contains the indices of the $k$ highest rated tokens. Fragment removal or retention operations are implemented consistently with the model pipeline by replacing selected positions with the PAD symbol, which preserves the sequence length and input type
as seen during training. Let $f(x) \in [0, 1]$ denote the probability of a positive class returned by the model.

The first metric is completeness. It assumes that the tokens indicated by the explanation actually carry information relevant to decision making, and after their removal, the model's confidence should decrease significantly. It is measured by the difference defined by the formula (\ref{eq1}).

\begin{equation}
	\Delta_{comp}(x, S_k) = f(x) - f(x \backslash S_k)
	\label{eq1}
\end{equation}

where $x \backslash Sk$ is a version of sequence $x$ with tokens from $S_k$ replaced by PAD. The larger $\Delta_{comp}$, the more accurately the explanation identified the fragments supporting the decision. 

The next metric is sufficiency, which reverses the perspective and assumes that the identified tokens alone are sufficient to support the model's decision. A sequence is created in which, except for $S_k$, all items are replaced by PAD, and the result is compared to the original.

\begin{equation}
	\Delta_{suff}(x, S_k) = f(x) - f(x \mid S_k)
	\label{eq2}
\end{equation}

A small value of $\Delta_{suff}$ indicates that the core information needed for the decision has been captured in $Top-k$.

The next measure is AOPC (emphArea Over the Perturbation Curve), which is the average loss of confidence when successively removing tokens in order of decreasing $|a_i|$. We construct a path
$x(0)=x, x(1), \ldots , x(m)$ (usually $m = k$), write $p(i) = f(x(i))$, and then compute

\begin{equation}
AOPC(x) = \frac{1}{m} \sum_{i=1}^m (p^{(0)} - p^{(i)})
\end{equation}

A high value means the model quickly loses the basis for its decision as subsequent top items are removed, so the ranking is faithful. Additionally, $Flip@k$ is calculated, which is the minimum number of top tokens from the ranking whose removal changes the label at a threshold of 0.5:

\begin{equation}
	\begin{aligned}
		Flip@k(x) = \min \Big\{ i \leqslant k : \, & 1\left [ f\left ( x^{(i)} \right ) \geqslant 0.5 \right ] \\
		& \neq 1\left [ f\left ( x^{(0)} \right ) \geqslant 0.5 \right ] \Big\}
	\end{aligned}
\end{equation}

Small values indicate that the explanation identifies the core of the decision, and only a few words are sufficient to reverse explain the prediction. Finally, the computational cost (average time) is also provided to provide a complete picture. This allows us to assess the usefulness of the methods in practical settings.

\section{Results of numerical experiments}

Tables \ref{res_lstm} and \ref{res_cnn} show a comparison of the explanation methods for both architectures (LSTM and CNN) using the four previously mentioned metrics.

\begin{table}[ht]
	\centering
	\caption{Quality of explanations for LSTM}\label{res_lstm}
	\begin{tabular}{l c c c c c}
	\hline
	\textbf{Method} & $\Delta_{comp}$ &  
	$\Delta_{suff}$ & \textbf{AOPC} & 
	$Flip@k$ & \textbf{Time [s]} \\
	\hline
	\textbf{IG} & 0.013654 & 0.502953 & 0.441086 & 9.850000 & 24.341803 \\
	\textbf{LIME} & 0.006688 & 0.508200 & 0.440015 & 9.650000 & 4.448272 \\
	\textbf{SHAP} & 0.086178 & 0.490652 & 0.472545 & 9.466667 & 4.693331 \\
	\hline
	\end{tabular}
\end{table}

For LSTM, SHAP delivers the best results. It has the strongest removal effect and the highest AOPC, while also having the lowest sufficiency. LIME is close, though it falls short in completeness. IG in this configuration performs worse in quality and is definitely the slowest, confirming that for a recursive architecture, the importance ranking with SHAP better follows the information aggregation method.

\begin{table}[ht]
	\centering
	\caption{Quality of explanations for CNN}\label{res_cnn}
	\begin{tabular}{l c c c c c}
		\hline
	\textbf{Method} & $\Delta_{comp}$ &  
	$\Delta_{suff}$ & \textbf{AOPC} & 
	$Flip@k$ & \textbf{Time [s]}  \\

		\hline
		\textbf{IG} & 0.286618 & 0.200787 & 0.649783 & 4.783333 & 0.155877 \\
		\textbf{LIME} & 0.173371 & 0.225614 & 0.615995 & 4.950000 & 0.736505 \\
		\textbf{SHAP} & 0.127419 & 0.399285 & 0.528378 & 4.933333 & 1.252607 \\
		\hline
	\end{tabular}
\end{table}

For CNNs, IG is the best method. It combines the highest completeness and AOPC with the lowest sufficiency and the lowest $Flip@k$, while also being very fast. LIME occupies an intermediate position, while SHAP primarily loses in sufficiency and AOPC. A lower $Flip@k$ than LSTM indicates the more localized nature of relevant fragments (a few key n-grams are sufficient to reverse the decision).

\section{Conclusions}
Two main interpretation issues emerge. First, within a single model, all explanation methods perform similarly: they differ directionally but not by large margins. This indicates that each method captures meaningful linguistic signal, so the choice can depend on computational cost and presentation needs (e.g., SHAP’s interactivity, LIME’s simplicity, IG’s efficiency). Second, XAI scores should not be directly compared between LSTM and CNN models, since they depend on classifier properties. Conclusions should therefore be drawn within each model, while crossmodel differences should be seen qualitatively as differences in how each architecture uses context, rather than as numeric comparisons.

Overall, the experiments show that strict, metricbased comparison of explanation methods is difficult, and their differences within a model are moderate. All techniques capture the core decision signal to a comparable degree, so selecting one should depend on analytical goals and practical limitations, not on a single best score.

Functionally, each method has a distinct profile: SHAP best reflects model architecture and produces clear textlevel explanations; Integrated Gradients is fast and lightweight; LIME offers quick surrogate models but captures internal model dynamics less precisely. Despite quantitative differences, all XAI methods add value by increasing transparency, revealing artifacts, supporting error analysis, improving stakeholder communication, and informing model design. In fakenews detection, this enhances risk assessment and trustworthiness beyond accuracy metrics.

The results also show that explanation quality depends strongly on model architecture: SHAP works best for LSTMs, while Integrated Gradients outperforms others for CNNs. This confirms that no XAI method is universally optimal; effectiveness is architecture dependent.

The adopted fidelity assessment procedure is also a limitation. All metrics were based on input perturbations implemented by replacing tokens with PAD. While this solution ensures technical consistency, it can lead to unnatural examples and distort the behavior of models, especially LSTMs, which accumulate context over time. The obtained metric values ??should therefore be interpreted in terms of relative comparisons rather than absolute quality measures.



Finally, it is important to emphasize the limitations of interpretation. The explanations obtained in this study are local in nature and cannot be generalized to the global performance of the model. While SHAP or LIME visualizations facilitated the analysis of specific examples, they did not provide a complete picture of the classifier's decisions across the entire data space. Furthermore, the observed differences in the attribution distribution between LSTM and CNN are largely due to the nature of the architectures, not solely to the performance of the explanation methods themselves. Failure to consider this aspect would risk overinterpreting the obtained importance maps.

%
%
%
%
%
\bibliographystyle{unsrt} 

\bibliography{biblio_xai2026} 

%
%
%
%
\end{document}